\documentclass[10pt,twocolumn,letterpaper]{article}
\pdfoutput=1 
\usepackage{cvpr}
\usepackage{times}
\usepackage{epsfig}
\usepackage{graphicx}
\usepackage{amsmath}
\usepackage{amssymb}

\usepackage{pdfpages}
\usepackage{xcolor}
\usepackage{makecell}
\usepackage{subcaption}
\usepackage{array,multirow}
\usepackage{booktabs}
\usepackage{pifont}
\usepackage[export]{adjustbox}
\usepackage{arydshln}

\usepackage[pagebackref=true,breaklinks=true,letterpaper=true,colorlinks,bookmarks=false]{hyperref}
\newcommand\norm[1]{\left\lVert#1\right\rVert}

\newcommand\blfootnote[1]{%
  \begingroup
  \renewcommand\thefootnote{}\footnote{#1}%
  \addtocounter{footnote}{-1}%
  \endgroup
}

\cvprfinalcopy 
\def\cvprPaperID{1500} 

\ifcvprfinal\fi
\begin{document}

\title{Relational Knowledge Distillation}

\author{Wonpyo Park\textsuperscript{*}\\
POSTECH\\
\and
Dongju Kim\\
POSTECH\\
\and
Yan Lu\\
Microsoft Research\\
\and
Minsu Cho\\
POSTECH\\
\and
{\tt\small \href{http://cvlab.postech.ac.kr/research/RKD/}{\textcolor{blue}{http://cvlab.postech.ac.kr/research/RKD/}}}
}

\maketitle

\blfootnote{\textsuperscript{*}The work was done when Wonpyo Park was an intern at MSR.}


\begin{abstract}
Knowledge distillation aims at transferring knowledge acquired in one model (a teacher) to another model (a student) that is typically smaller.
Previous approaches can be expressed as a form of training the student to mimic output activations of individual data examples represented by the teacher.
We introduce a novel approach, dubbed relational knowledge distillation (RKD), that transfers mutual relations of data examples instead.
For concrete realizations of RKD, we propose distance-wise and angle-wise distillation losses that penalize structural differences in relations.
Experiments conducted on different tasks show that the proposed method improves educated student models with a significant margin.
In particular for metric learning, it allows students to outperform their teachers' performance, achieving the state of the arts on standard benchmark datasets.
\end{abstract}

\section{Introduction}

Recent advances in computer vision and artificial intelligence have largely been driven by deep neural networks with many layers, and thus current state-of-the-art models typically require a high cost of computation and memory in inference. 
One promising direction for mitigating this computational burden is to transfer {\em knowledge} in the cumbersome model (a teacher) into a small model (a student). 
To this end, there exist two main questions: (1) `what constitutes the knowledge in a learned model?' and (2) `how to transfer the knowledge into another model?'. Knowledge distillation (or transfer) (KD) methods~\cite{breiman1996born,bucilua2006compression,hinton2015distilling} assume the knowledge as a learned mapping from inputs to outputs, and transfer the knowledge by training the student model with the teacher's outputs (of the last or a hidden layer) as targets. Recently, KD has turned out to be very effective not only in training a student model~\cite{NIPS2014_5484, hinton2015distilling,huang2017like,romero2014fitnets,zagoruyko2016paying}  but also in improving a teacher model itself by self-distillation~\cite{bagherinezhad2018label,furlanello2018born,Yim2017AGF}.


\begin{figure}
\centering
\scalebox{0.45}{\includegraphics[width=\textwidth]{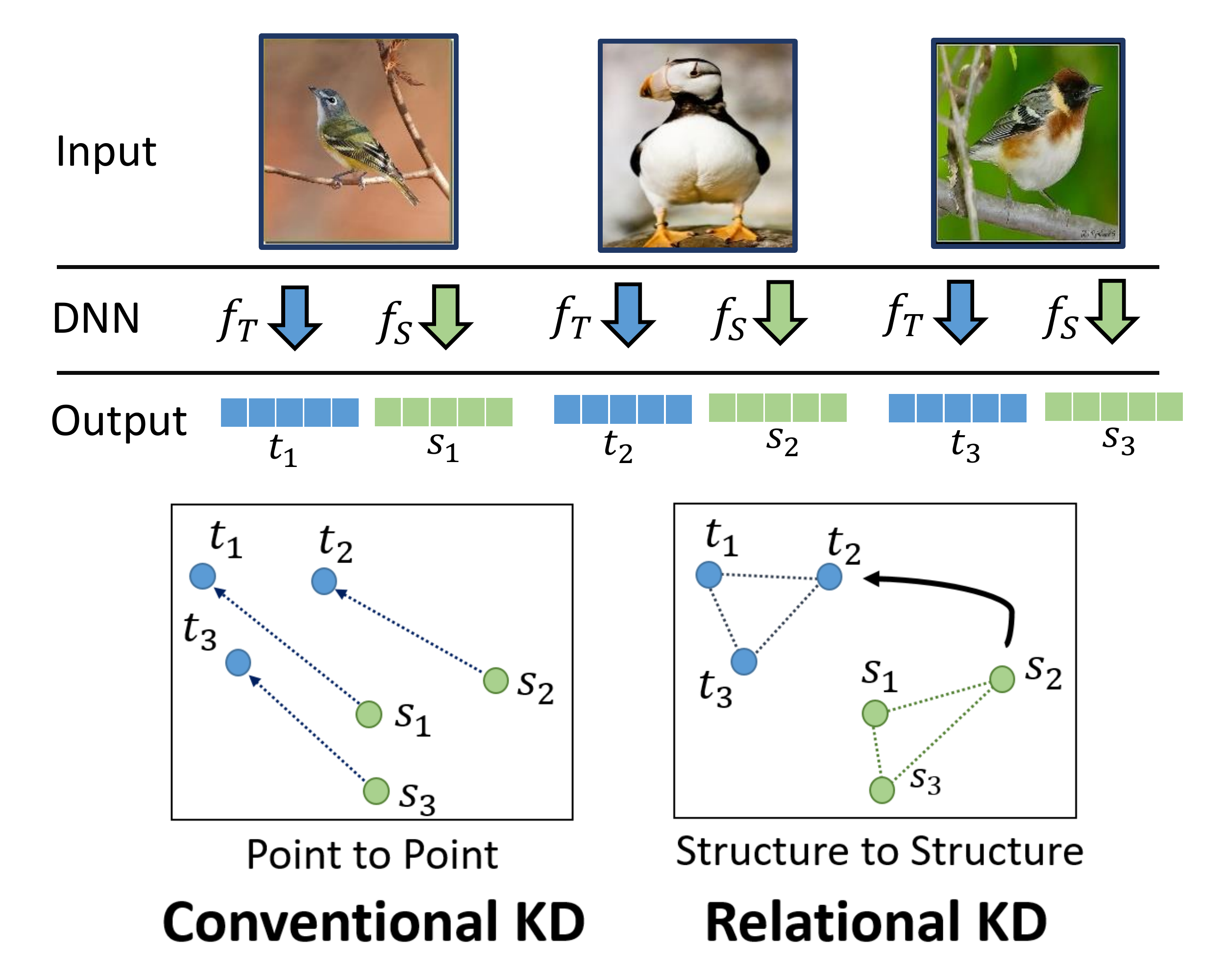}}
\caption{Relational Knowledge Distillation. While conventional KD transfers individual outputs from a teacher model ($f_T$) to a student model ($f_S$) point-wise, our approach transfers relations of the outputs structure-wise. It can be viewed as a generalization of conventional KD.}\label{fig:rkd_teaser}\vspace{-2mm}
\vspace{-0.3cm}
\end{figure}

In this work, we revisit KD from a perspective of the linguistic structuralism~\cite{matthews2001short}, which focuses on structural relations in a semiological system. 
Saussure's concept of the relational identity of signs is at the heart of structuralist theory; ``In a language, as in every other semiological system, what distinguishes a sign is what constitutes it''~\cite{saussure1983course}.
In this perspective, the meaning of a sign depends on its relations with other signs within the system; a sign has no absolute meaning independent of the context. 

The central tenet of our work is that what constitutes the knowledge is better presented by {\em relations} of the learned representations than individuals of those; an individual data example, \eg, an image, obtains a meaning in relation to or in contrast with other data examples in a system of representation, and thus primary information lies in the structure in the data embedding space. 
On this basis, we introduce a novel approach to KD, dubbed \textit{Relational Knowledge Distillation} (RKD), that transfers structural relations of outputs rather than individual outputs themselves (Figure~\ref{fig:rkd_teaser}). 
For its concrete realizations, we propose two RKD losses: distance-wise (second-order) and angle-wise (third-order) distillation losses. RKD can be viewed as a generalization of conventional KD, and also be combined with other methods to boost the performance due to its complementarity with conventional KD.
In experiments on metric learning, image classification, and few-shot learning, our approach significantly improves the performance of student models.  
Extensive experiments on the three different tasks show that knowledge lives in the relations indeed, and RKD is effective in transferring the knowledge.


\section{Related Work}

    

There has been a long line of research and development on transferring knowledge from one model to another. Breiman and Shang~\cite{breiman1996born} first proposed to learn single-tree models that approximate the performance of multiple-tree models and provide better interpretability. 
Similar approaches for neural networks have been emerged in the work of Bucilua \etal.~\cite{bucilua2006compression}, Ba and Caruana~\cite{NIPS2014_5484}, and Hinton \etal.~\cite{hinton2015distilling}, mainly for the purpose of model compression. 
Bucilua~\etal compress an ensemble of neural networks into a single neural network. Ba and Caruana~\cite{NIPS2014_5484} increase the accuracy of a shallow neural network by training it to mimic a deep neural network with penalizing the difference of logits between the two networks. 
Hinton \etal~\cite{hinton2015distilling} revive this idea under the name of KD that trains a student model with the objective of matching the softmax distribution of a teacher model. Recently, many subsequent papers have proposed different approaches to KD.
Romero \etal~\cite{romero2014fitnets} distill a teacher using additional linear projection layers to train a relatively narrower students. Instead of imitating output activations of the teacher, Zagoruyko and Komodakis~\cite{zagoruyko2016paying} and Huang and Wang~\cite{huang2017like} transfer an attention map of a teacher network into a student, and Tarvainen and Valpola ~\cite{DBLP:journals/corr/TarvainenV17} introduce a similar approach using mean weights. Sau \etal~\cite{sau2016deep} propose a noise-based regularizer for KD while Lopes \etal~\cite{DBLP:journals/corr/abs-1710-07535} introduce data-free KD that utilizes metadata of a teacher model.
Xu \etal~\cite{xu2018training} propose a conditional adversarial network to learn a loss function for KD.
Crowley \etal \cite{crowley2017moonshine} compress a model by grouping convolution channels of the model and training it with an attention transfer. Polino \etal \cite{polino2018model} and Mishra and Marr \cite{mishra2018apprentice} combine KD with network quantization, which aims to reduce bit precision of weights and activations.  


    

A few recent papers \cite{bagherinezhad2018label,furlanello2018born,Yim2017AGF} have shown that distilling a teacher model into a student model of identical architecture, \ie, self-distillation, can improve the student over the teacher. Furlanello \etal \cite{furlanello2018born} and Bagherinezhad \etal \cite{bagherinezhad2018label} demonstrate it by training the student using softmax outputs of the teacher as ground truth over generations. Yim \etal ~\cite{Yim2017AGF} transfers output activations using Gramian matrices and then fine-tune the student. We also demonstrate that RKD strongly benefits from self-distillation. 

KD has also been investigated beyond supervised learning.
Lopez-Paz \etal \cite{LopSchBotVap16} unify two frameworks \cite{hinton2015distilling, Vapnik:2015:LUP:2789272.2886814} and extend it to unsupervised, semi-supervised, and multi-task learning scenarios. 
Radosavovic~\etal~\cite{radosavovic2017dd} generate multiple predictions from  an example by applying multiple data transformations on it, then use an ensemble of the predictions as annotations for omni-supervised learning.
    
With growing interests in KD, task-specific KD methods have been proposed for object detection \cite{8268087,Chen2017LearningEO,46626}, face model compression~\cite{luo2016mobileid}, and image retrieval and Re-ID~\cite{chen2017darkrank}.
Notably, the work of Chen~\etal~\cite{chen2017darkrank} proposes a KD technique for metric learning that transfers similarities between images using a rank loss. 
In the sense that they transfer relational information of ranks, it has some similarity with ours.
Their work, however, is only limited to metric learning whereas we introduce a general framework for RKD and demonstrate its applicability to various tasks. Furthermore, our experiments on metric learning show that the proposed method outperforms  \cite{chen2017darkrank} with a significant margin.



\section{Our Approach}

\begin{figure*}
\includegraphics[width=\textwidth]{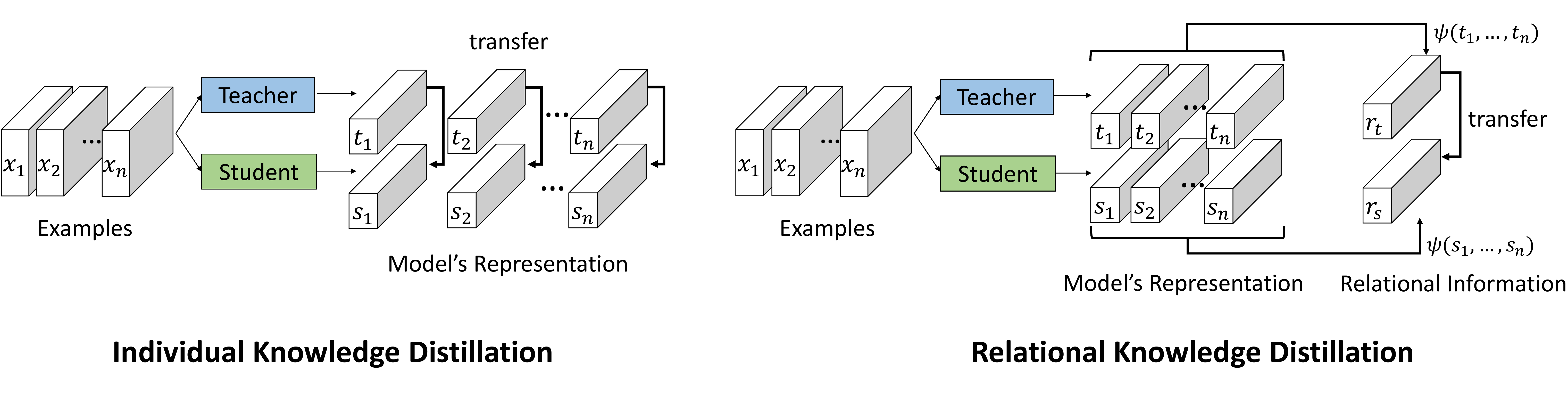}
\centering
\vspace{-1cm}
\caption{Individual knowledge distillation (IKD) vs. relational knowledge distillation (RKD). While conventional KD (IKD) transfers individual outputs of the teacher directly to the student, RKD extracts relational information using a relational potential function $\psi(\cdot)$, and transfers the information from the teacher to the student.}\label{fig:individual_relational_distillation}
\vspace{-0.3cm}
\end{figure*}

In this section we first revisit conventional KD and introduce a general form of RKD. Then, two simple yet effective distillation losses will be proposed as instances of RKD. 

\bigbreak
\noindent\textbf{Notation.} Given a teacher model $T$ and a student model $S$, we let $f_T$ and $f_S$ be functions of the teacher and the student, respectively. 
Typically the models are deep neural networks and in principle the function $f$ can be defined using output of any layer of the network (\eg, a hidden or softmax layer). 
We denote by $\mathcal{X}^N$ a set of $N$-tuples of distinct data examples, \eg, $\mathcal{X}^2 = \{(x_i, x_j) | i \neq j\}$ and $\mathcal{X}^3 = \{(x_i, x_j, x_k) | i \neq j \neq k\}$.

\subsection{Conventional knowledge distillation}
In general, conventional KD methods~\cite{NIPS2014_5484,bagherinezhad2018label,crowley2017moonshine,hinton2015distilling,huang2017like,polino2018model,romero2014fitnets,Yim2017AGF,zagoruyko2016paying}
can commonly be expressed as minimizing the objective function:  
\begin{align}
\label{eq:point_distillation}
\mathcal{L}_{\text{IKD}} =  \sum_{x_i \in \mathcal{X}} l\big(f_T(x_i), f_S(x_i)\big), 
\end{align}
where $l$ is a loss function that penalizes the difference between the teacher and the student. 

For example, the popular work of Hinton \etal \cite{hinton2015distilling} uses pre-softmax outputs for $f_T$ and $f_S$, and puts softmax (with temperature $\tau$) and Kullback-Leibler divergence for $l$: 
\begin{align}
\label{eq:dark_knowledge_distillation}
\sum_{x_i \in \mathcal{X}} \mathrm{KL}\Big(\text{softmax}\big(\frac{f_T(x_i)}{\tau}\big), \text{softmax}\big(\frac{f_S(x_i)}{\tau}\big)\Big). 
\end{align}
The work of Romero \etal \cite{romero2014fitnets} propagates knowledge of hidden activations by setting $f_T$ and $f_S$ to be outputs of hidden layers, and $l$ to be squared Euclidean distance. As the hidden layer output of the student usually has a smaller dimension than that of the teacher, a linear mapping $\beta$ is introduced to bridge the different dimensions:
\begin{align}
\label{eq:fitnet}
\sum_{x_i \in \mathcal{X}} \norm{f_T(x_i) - \beta\big(f_S(x_i)\big)}_{2}^2. 
\end{align}

Likewise, many other methods \cite{NIPS2014_5484,bagherinezhad2018label, crowley2017moonshine,huang2017like,polino2018model,Yim2017AGF,zagoruyko2016paying} can also be formulated as a form of Eq.~(\ref{eq:point_distillation}). 
Essentially, conventional KD transfers {\em individual} outputs of the teacher to the student.
We thus call this category of KD methods as \textit{Individual KD} (IKD).


\subsection{Relational knowledge distillation}
\label{method:rkd}

RKD aims at transferring structural knowledge using mutual relations of data examples in the teacher's output presentation. 
Unlike conventional approaches, it computes a relational potential $\psi$ for each $n$-tuple of data examples and transfers information through the potential from the teacher to  the student. 

For notational simplicity, let us define $t_i=f_T(x_i)$ and $s_i=f_S(x_i)$. The objective for RKD is expressed as 
\begin{align}
   \label{eq:relational_knowledge_distillation}
   \mathcal{L}_{\text{RKD}} & = \sum_{(x_1, ..,x_n) \in \mathcal{X}^N}l\big(\psi(t_1, .., t_n), \psi(s_1, ..,s_n)\big), 
\end{align}
where $(x_1,x_2, ...,x_n)$ is a $n$-tuple drawn from $\mathcal{X}^N$, $\psi$ is a relational potential function that measures a relational energy of the given $n$-tuple, and $l$ is a loss that penalizes difference between the teacher and the student. RKD trains the student model to form the same relational structure with that of the teacher in terms of the relational potential function used. Thanks to the  potential, it is able to transfer knowledge of high-order properties, which is invariant to lower-order properties, even regardless of difference in output dimensions between the teacher and the student. 
RKD can be viewed as a generalization of IKD in the sense that Eq.~(\ref{eq:relational_knowledge_distillation}) above reduces to Eq.~(\ref{eq:point_distillation}) when the relation is unary ($N=1$)  and the potential function $\psi$ is identity. Figure~\ref{fig:individual_relational_distillation} illustrates comparison between IKD and RKD. 

As expected, the relational potential function $\psi$ plays a crucial role in RKD; the effectiveness and efficiency of RKD relies on the choice of the potential function. For example, a higher-order potential may be powerful in capturing a higher-level structure but be more expensive in computation. 
In this work, we propose two simple yet effective potential functions and corresponding losses for RKD, which exploit pairwise and ternary relations of examples, respectively: \textit{distance-wise} and \textit{angle-wise} losses.

\vspace{-0.3cm}
\subsubsection{Distance-wise distillation loss}
Given a pair of training examples, \textit{distance-wise} potential function $\psi_\mathrm{D}$ measures the Euclidean distance between the two examples in the output representation space: 
\begin{align}
    \label{eq:rel-distance-wise}
 \psi_\text{D}(t_i, t_j \space)  = \frac{1}{\mu} \norm{t_i - t_j}_2, 
\end{align}
where $\mu$ is a normalization factor for distance. 
To focus on relative distances among other pairs, we set $\mu$ to be the average distance between pairs from $\mathcal{X}^2$ in the mini-batch:   
\begin{align}
 \mu = \frac{1}{|\mathcal{X}^2|} \sum_{(x_i, x_j) \in \mathcal{X}^2} \norm{t_i - t_j}_2. 
\end{align}
Since distillation attempts to match the distance-wise potentials between the teacher and the student, this mini-batch distance normalization is useful particularly when there is a significant difference in scales between teacher distances $\norm{t_i - t_j}_2$  and student distances $\norm{s_i - s_j}_2$, \eg, due to the difference in output dimensions.
In our experiments, we observed that the normalization provides more stable and faster convergence in training. 

Using the distance-wise potentials measured in both the teacher and the student, a distance-wise distillation loss is defined as
\begin{align}
\mathcal{L}_{\text{RKD-D}}=\sum_{(x_i, x_j) \in \mathcal{X}^2} l_{\delta}\big(\psi_{\text{D}}(t_i, t_j), \psi_{\text{D}}(s_i, s_j)\big), 
\end{align}
where $l_{\delta}$ is Huber loss, which is defined as  
\begin{align}
    \label{eq:huber_loss}
    l_\delta (x, y) = \begin{cases}
     \frac{1}{2}{(x-y)^2}                  & \text{for } |x-y| \le 1, \\
     |x-y| - \frac{1}{2}, & \text{otherwise.}
    \end{cases}
\end{align}

The distance-wise distillation loss transfers the relationship of examples by penalizing distance differences between their output representation spaces. Unlike conventional KD, it does not force the student to match the teacher's output directly, but encourages the student to focus on distance structures of the outputs. 


\vspace{-0.3cm}
\subsubsection{Angle-wise distillation loss}

Given a triplet of examples, an angle-wise relational potential measures the angle formed by the three examples in the output representation space: 
\begin{align}
& \psi_{\text{A}}(t_i, t_j, t_k) = \cos\angle t_i t_j t_k= {\langle \mathbf{e}^{ij}, \mathbf{e}^{kj} \rangle} \\ \nonumber 
& \text{where} \quad \mathbf{e}^{ij} = \frac{t_i - t_j}{\norm{t_i - t_j}_2},  \mathbf{e}^{kj} = \frac{t_k - t_j}{\norm{t_k - t_j}_2}.
\end{align}
Using the angle-wise potentials measured in both the teacher and the student, an angle-wise distillation loss is defined as 
\begin{align}
    \mathcal{L}_{\text{RKD-A}} = \sum_{(x_i,x_j,x_k) \in \mathcal{X}^3} l_{\delta}\big(\psi_{\text{A}}(t_i, t_j, t_k), \psi_{\text{A}}(s_i, s_j, s_k)\big), 
\end{align}
where $l_{\delta}$ is the Huber loss. The \textit{angle-wise} distillation loss transfers the relationship of training example embeddings by penalizing angular differences. Since an angle is a higher-order property than a distance, it may be able to transfer relational information more effectively, giving more flexibility to the student in training. In our experiments, we observed that the angle-wise loss often allows for faster convergence and better performance.

\vspace{-0.3cm}
\subsubsection{Training with RKD}
During training, multiple distillation loss functions, including the proposed RKD losses, can be used either alone or together with task-specific loss functions, \eg, cross-entropy for classification. 
Therefore, the overall objective has a form of 
\begin{align}
    \label{eq:train_loss}
    \mathcal{L}_\mathrm{task} + \lambda_\text{KD} \cdot \mathcal{L}_\mathrm{KD}, 
\end{align}
where $\mathcal{L}_\mathrm{task}$ is a task-specific loss for the task at hand, $\mathcal{L}_\mathrm{KD}$ is a knowledge distillation loss, and $\lambda_\text{KD}$ is a tunable hyperparameter to balance the loss terms. 
When multiple KD losses are used during training, each loss is weighted with a corresponding balancing factor.
In sampling tuples of examples for the proposed distillation losses, 
we simply use all possible tuples (\ie, pairs or triplets) from examples in a given mini-batch. 

\vspace{-0.3cm}
\subsubsection{Distillation target layer}

For RKD, the distillation target function $f$ can be chosen as output of any layer of teacher/student networks in principle. However, since the distance/angle-wise losses do not transfer individual outputs of the teacher, it is not adequate to use them {\em alone} to where the individual output values themselves are crucial, \eg, softmax layer for classification. In that case, it needs to be used together with IKD losses or task-specific losses. In most of the other cases, RKD is applicable and effective in our experience. We demonstrate its efficacy in the following section.

\section{Experiments}

We evaluate RKD on three different tasks: metric learning, classification, and few-shot learning. 
Throughout this section, we refer to RKD with the distance-wise loss as RKD-D, that with angle-wise loss as RKD-A, and that with two losses together as RKD-DA. When the proposed losses are combined with other losses during training, we assign respective balancing factors to the loss terms. 
We compare RKD with other KD methods, \eg, FitNet~\cite{romero2014fitnets}\footnote{When FitNet is used, following the original paper, we train the model with two stages: (1) train the model with FitNet loss, and (2) fine-tune the model with the task-specific loss at hand.}, Attention~\cite{zagoruyko2016paying} and HKD (Hinton's KD)~\cite{hinton2015distilling}.
For metric learning, we conduct an additional comparison with DarkRank~\cite{chen2017darkrank} which is a KD method specifically designed for metric learning.
For fair comparisons, we tune hyperparameters of the competing methods using grid search.

Our code used for experiments is available online: \href{http://cvlab.postech.ac.kr/research/RKD/}{\textcolor{blue}{http://cvlab.postech.ac.kr/research/RKD/}}.

\subsection{Metric learning}

We first evaluate the proposed method on metric learning where relational knowledge between data examples appears to be most relevant among other tasks.
Metric learning aims to train an embedding model that projects data examples onto a manifold where two examples are close to each other if they are semantically similar and otherwise far apart.
As embedding models are commonly evaluated on image retrieval, we validate our approach using image retrieval benchmarks of CUB-200-2011~\cite{WahCUB_200_2011}, Cars 196~\cite{KrauseStarkDengFei-Fei_3DRR2013}, and Stanford Online Products~\cite{Song2016DeepML} datasets and we follow the train/test splits suggested in~\cite{Song2016DeepML}.
For the details of the datasets, we refer the readers to the corresponding papers. 

For an evaluation metric, recall@K is used. Once all test images are embedded using a model, 
each test image is used as a query and top K nearest neighbor images are retrieved from the test set excluding the query.
Recall for the query is considered 1 if the retrieved images contain the same category with the query. 
Recall@K are computed by taking the average recall over the whole test set.

\begin{table*}[t!]
    \centering
    \caption{Recall@1 on CUB-200-2011 and Cars 196. The teacher is based on ResNet50-512. Model-\textit{d} refers to a network with \textit{d} dimensional embedding. `O' indicates models trained with $\ell2$ normalization, while `X' represents ones without it.}
    \label{tab:recall_small_dim}
      
 
   \begin{subtable}{\linewidth}\centering
    \caption{Results on CUB-200-2011 \cite{WahCUB_200_2011}}
    \label{table:1a}
    \small
    \begin{tabular}{l|c|c|c|c|ccc}
      & \multirow{2}{*}{\shortstack{Baseline\\(Triplet~\cite{DBLP:journals/corr/SchroffKP15})}}
      & \multirow{2}{*}{FitNet~\cite{romero2014fitnets}} 
      & \multirow{2}{*}{Attention~\cite{zagoruyko2016paying}}
      & \multirow{2}{*}{DarkRank~\cite{chen2017darkrank}}
      & \multicolumn{3}{c}{Ours}  \\ 
      & & & & & \small RKD-D & RKD-A & RKD-DA \\
      \Xhline{2\arrayrulewidth}
      $\ell2$ normalization & O & O & O & O &  O\quad /\quad X &  O\quad /\quad X  & O\quad /\quad X  \\
      \hline
      ResNet18-16 & 37.71 & 42.74 & 37.68 & 46.84  & 46.34 / 48.09 & 45.59 / \textbf{48.60} & 45.76 / 48.14\\
      ResNet18-32 & 44.62 & 48.60 & 45.37 & 53.53  & 52.68 / \textbf{55.72} & 53.43 / 55.15 & 53.58 / 54.88\\
      ResNet18-64 & 51.55 & 51.92 & 50.81 & 56.30  & 56.92 / 58.27 & 56.77 / 58.44 & 57.01 / \textbf{58.68}\\
      ResNet18-128 & 53.92 & 54.52 & 55.03 & 57.17 & 58.31 / 60.31 & 58.41 / \textbf{60.92} & 59.69 / 60.67\\
        \hline
        \hline
        ResNet50-512 & 61.24 & \multicolumn{5}{c}{} \\
      
    \end{tabular}   
 \end{subtable}
 
     \small

 
 \begin{subtable}{\linewidth}\centering
    \caption{Results on Cars 196 \cite{KrauseStarkDengFei-Fei_3DRR2013}}
    \label{table:1b}
    \begin{tabular}{l|c|c|c|c|ccc}
      & \multirow{2}{*}{\shortstack{Baseline\\(Triplet~\cite{DBLP:journals/corr/SchroffKP15})}}
      & \multirow{2}{*}{FitNet~\cite{romero2014fitnets}} 
      & \multirow{2}{*}{Attention~\cite{zagoruyko2016paying}}
      & \multirow{2}{*}{DarkRank~\cite{chen2017darkrank}}
      & \multicolumn{3}{c}{Ours}  \\ 
      & & & & & \small RKD-D & RKD-A & RKD-DA \\
      \Xhline{2\arrayrulewidth}
      $\ell2$ normalization & O & O & O & O & O\quad /\quad X &  O\quad /\quad X  & O\quad /\quad X  \\
      \hline
      ResNet18-16 & 45.39 & 57.46 & 46.44 & 64.00 & 63.23 / 66.02 & 61.39 / \textbf{66.25} & 61.78 / 66.04\\
      ResNet18-32 & 56.01 & 65.81 & 59.40 & 72.41 & 73.50 / \textbf{76.15} & 73.23 / 75.89 & 73.12 / 74.80\\
      ResNet18-64 & 64.53 & 70.67 & 67.24 & 76.20 & 78.64 / \textbf{80.57} & 77.92 / 80.32 & 78.48 / 80.17\\
      ResNet18-128 & 68.79 & 73.10 & 71.95 & 77.00 & 79.72 / 81.70 & 80.54 / 82.27 & 80.00 / \textbf{82.50}\\
        \hline
        \hline
        ResNet50-512 & 77.17 & \multicolumn{5}{c}{}  \\
    \end{tabular}   
 \end{subtable}
\end{table*}

For training, we follows the protocol of \cite{wu2017sampling}. We obtain training samples by randomly cropping  $224\times224$ images from resized $256\times256$ images and applying random horizontal flipping for data augmentation. During evaluation, we use a single center crop. All models are trained using Adam optimizer with batch size of 128 for all datasets. 
For effective pairing, we follow batch construction from FaceNet \cite{DBLP:journals/corr/SchroffKP15}, and sample 5 positive images per category in a mini-batch.

For a teacher model, ResNet50 \cite{DBLP:journals/corr/HeZRS15}, which is pre-trained on ImageNet ILSVRC dataset \cite{DBLP:journals/corr/RussakovskyDSKSMHKKBBF14}, is used. We take layers of the network upto \texttt{avgpool} and append a single fully-connected layer with embedding size of 512 followed by $\ell2$ normalization. For a student model, ResNet18 \cite{DBLP:journals/corr/HeZRS15}, which is also ImageNet-pretrained, is used in a similar manner but with different embedding sizes.
The teacher models are trained with the triplet loss \cite{DBLP:journals/corr/SchroffKP15}, which is the most common and also effective in metric learning.

\smallbreak
\noindent\textbf{Triplet}~\cite{DBLP:journals/corr/SchroffKP15}. 
When an anchor $x_a$, positive $x_p$ and negative $x_n$ are given, the triplet loss enforces the squared euclidean distance between anchor and negative to be larger than that between anchor and positive by margin $m$: 
\begin{align}
    \mathcal{L}_{\text{triplet}} = \left[\norm{f(x_a) - f(x_p)}_{2}^2 - \norm{f(x_a) - f(x_n)}_{2}^2 + m\right]_{+}.
\end{align}
We set the margin $m$ to be 0.2 and use distance-weighted sampling \cite{wu2017sampling} for triplets.
We apply $\ell2$ normalization at the final embedding layer such that the embedding vectors have a unit length, \ie, $\norm{f(x)}$=1.
Using $\ell2$ normalization is known to stabilize training of the triplet loss by restricting the range of distances between embedding points to [0, 2].
Note that $\ell2$ normalization for embedding is widely used in deep metric learning~\cite{chen2017darkrank,Kim_2018_ECCV,Song2016DeepML,opitz_2018_pami,NIPS2016_6200,Wang_2017_ICCV}. 

\smallbreak
\noindent\textbf{RKD}. We apply RKD-D and RKD-A on the final embedding outputs of the teacher and the student.
Unlike the triplet loss, the proposed RKD losses are not affected by range of distance between embedding points, and do not have sensitive hyperparameters to optimize such as margin $m$ and triplet sampling parameters. 
To show the robustness of RKD, we compare RKD {\em without} $\ell2$ normalization to RKD {\em with} $\ell2$ normalization.
For RKD-DA, we set $\lambda_\text{RKD-D} = 1$ and $\lambda_\text{RKD-A}=2$.
Note that for metric learning with RKD losses, we do not use the task loss, \ie, the triplet loss, so that the model is thus trained purely by teacher's guidance without original ground-truth labels; using the task loss does not give additional gains in our experiments. 

\smallbreak
\noindent\textbf{Attention}~\cite{zagoruyko2016paying}. Following the original paper, we apply the method on the output of the second, the third, and the fourth blocks of ResNet. We set $\lambda_\text{Triplet} = 1$ and $\lambda_\text{Attention} = 50$.

\smallbreak
\noindent\textbf{FitNet}~\cite{romero2014fitnets}. Following the original paper, we train a model in two stages; we first initialize a model with FitNet loss, and then fine-tune the model, in our case, with Triplet. We apply the method on outputs of the second, the third, and the fourth blocks of ResNet, as well as the final embedding. 

\smallbreak
\noindent\textbf{DarkRank}~\cite{chen2017darkrank} is a KD method for metric learning that transfers similarity ranks  between data examples. 
Among two losses proposed in  \cite{chen2017darkrank}, we use the HardRank loss as it is computationally efficient and also comparable to the other in performance. 
The DarkRank loss is applied on final outputs of the teacher and the student.
In training, we use the same objective with the triplet loss as suggested in the paper. We carefully tune hyperparameters of DarkRank to be optimal: $\alpha=3$, $\beta=3$, and $\lambda_\text{DarkRank} = 1$, and $\lambda_\text{Triplet} = 1$; we conduct a grid search on $\alpha$ (1 to 3), $\beta$ (2 to 4), $\lambda_\text{DarkRank}$ (1 to 4). In our experiment, our hyperparameters give better results than those used in \cite{chen2017darkrank}.

\vspace{-0.3cm}
\subsubsection{Distillation to smaller networks}
\label{sec:distill_to_small}

Table \ref{tab:recall_small_dim} shows image retrieval performance of student models with different embedding dimensions on CUB-200-2011 \cite{Vapnik:2015:LUP:2789272.2886814} and Cars 196 \cite{KrauseStarkDengFei-Fei_3DRR2013}.  
RKD significantly improves the performance of student networks compared to the baseline model, which is directly trained with Triplet, also outperforms DarkRank by a large margin.
Recall@1 of Triplet decreases dramatically with smaller embedding dimensions while that of RKD is less affected by embedding dimensions; the relative gain of recall@1 by RKD-DA increases from 12.5, 13.8, 23.0, to 27.7 on CUB-200-2011, and
from 20.0, 24.2, 33.5 to 45.5 on Cars 196.
The results also show that RKD benefits from training without $\ell2$ normalization by exploiting a larger embedding space. Note that the absence of $\ell2$ normalization has degraded all the other methods in our experiments. 
Surprisingly, by RKD on Cars 196, students with the smaller backbone and less embedding dimension even outperform their teacher, \eg, 77.17 of ResNet50-512 teacher vs. 82.50 of ResNet18-128 student. 


\vspace{-0.3cm}
\subsubsection{Self-distillation}

As we observe that RKD is able to improve smaller student models over its teacher, 
we now conduct self-distillation experiments where the student architecture is identical to the teacher architecture. 
Here, we do not apply $\ell2$ normalization on students to benefit from the effect as we observe in the previous experiment. 
The students are trained with RKD-DA over generations by using the student from the previous generation as a new teacher.
Table \ref{tab:teacher_to_teacher_metric} shows the result of self-distillation where `CUB', `Cars', and `SOP' refer to CUB-200-2011~\cite{WahCUB_200_2011}, Cars 196~\cite{KrauseStarkDengFei-Fei_3DRR2013}, and Stanford Online Products~\cite{Song2016DeepML}, respectively.
All models consistently outperform initial teacher models, which are trained with the triplet loss. 
In particular,  student models of CUB-200-2011 and Cars 196 outperform initial teachers with a significant gain.
However, the performances do not improve from the second generation in our experiments.

%

\begin{table}[h!]
  \begin{center}
    \caption{Recall@1 of self-distilled models. Student and teacher models have the same architecture.
    The model at Gen(\textit{n}) is guided by the model at Gen(\textit{n}-1).}
    \label{tab:teacher_to_teacher_metric}
    \small
    \begin{tabular}{l|c|c|c} 
       & CUB~\cite{WahCUB_200_2011} & Cars~\cite{KrauseStarkDengFei-Fei_3DRR2013} & SOP~\cite{Song2016DeepML} \\
      \hline
    ResNet50-512-Triplet & 61.24 & 77.17 & 76.58\\
    \hline
    ResNet50-512@Gen1 & \textbf{65.68} & \textbf{85.65} & \textbf{77.61} \\
    ResNet50-512@Gen2 & 65.11 & 85.61 & 77.36 \\
    ResNet50-512@Gen3 & 64.26 & 85.23 & 76.96 \\
    \end{tabular}
    \vspace{-0.7cm}
  \end{center}
\end{table}

\begin{table*}[t!p]
    \centering
    \caption{Recall@K comparison with state of the arts on CUB-200-2011, Car 196, and Stanford Online Products. We divide methods into two groups according to backbone networks used. A model-\textit{d} refers to model with \textit{d}-dimensional embedding. Boldfaces represent the best performing model for each backbone while underlines denote the best among all the models.}
    \label{tab:sota_metriclearning}
    \vspace{-0.2cm}
    \small
    \begin{tabular}{l|l|cccc|cccc|cccc}
      \multicolumn{2}{c}{} & \multicolumn{4}{c}{CUB-200-2011 \cite{WahCUB_200_2011}} 
       & \multicolumn{4}{c}{Cars 196 \cite{KrauseStarkDengFei-Fei_3DRR2013}}
       & \multicolumn{4}{c}{Stanford Online Products \cite{Song2016DeepML}} \\
      \Xhline{2\arrayrulewidth}
         & K & 1 & 2 & 4 & 8 & 1 & 2 & 4 & 8 & 1 & 10 & 100 & 1000  \\
      \Xhline{2\arrayrulewidth}
     \multirow{7}{*}{GoogLeNet~\cite{szegedy2015going}} & LiftedStruct~\cite{Song2016DeepML}-128 & 47.2 & 58.9 & 70.2 & 80.2 & 49.0 & 60.3 & 72.1 & 81.5 & 62.1 & 79.8 & 91.3 & 97.4\\
     & N-pairs~\cite{NIPS2016_6200}-64& 51.0 & 63.3 & 74.3 & 83.2 & 71.1 & 79.7 & 86.5 & 91.6 & 67.7 & 83.8 & 93.0 & 97.8\\
     & Angular~\cite{Wang_2017_ICCV}-512 & 54.7 & 66.3 & 76.0 & 83.9 & 71.4 & 81.4 & 87.5 & 92.1 & 70.9 & 85.0 & 93.5 & 98.0 \\
     & A-BIER~\cite{opitz_2018_pami}-512 & 57.5 & 68.7 & 78.3 & 86.2 & 82.0 & 89.0 & 93.2 & 96.1 & 74.2 & 86.9 & 94.0 & 97.8\\
     & ABE8~\cite{Kim_2018_ECCV}-512 &  60.6 & 71.5 & 79.8 & 87.4 & \underline{\textbf{85.2}} & \textbf{90.5} & 94.0 & 96.1 & \textbf{76.3} & \textbf{88.4} & 94.8 & 98.2\\
     & RKD-DA-128 & 60.8 & 72.1 &	81.2 &	\textbf{89.2} & 81.7 & 88.5 & 93.3 & 96.3 & 74.5 & 88.1 & 95.2 &	98.6\\
     & RKD-DA-512 &  \textbf{61.4} & \textbf{73.0} & \textbf{81.9} & 89.0 & 82.3 & 89.8 & \textbf{94.2} &	\textbf{96.6} & 75.1 & 88.3 & \textbf{95.2} & \textbf{98.7}\\
      \hline
      \hline
     \multirow{3}{*}{ResNet50~\cite{DBLP:journals/corr/HeZRS15}}  &  Margin~\cite{wu2017sampling}-128 & 63.6 & 74.4 & 83.1 & 90.0 & 79.6 & 86.5 & 91.9 & 95.1 & 72.7 & 86.2 & 93.8 & 98.0\\
     & RKD-DA-128 & \underline{\textbf{64.9}} & \underline{\textbf{76.7}} & \underline{\textbf{85.3}} & \underline{\textbf{91.0}} & \textbf{84.9} & \underline{\textbf{91.3}} & \underline{\textbf{94.8}} & \underline{\textbf{97.2}} & \underline{\textbf{77.5}} & \underline{\textbf{90.3}} & \underline{\textbf{96.4}} & \underline{\textbf{99.0}} \\

    \end{tabular}   
\end{table*}

\vspace{-0.3cm}
\subsubsection{Comparison with state-of-the art methods}

We compare the result of RKD with state-of-the art methods for metric learning.
Most of recent methods adopt GoogLeNet~\cite{szegedy2015going} as a backbone while the work of~\cite{wu2017sampling} uses a variant of ResNet50~\cite{DBLP:journals/corr/HeZRS15} with a modified number of channels. 
For fair comparisons, we train student models on both GoogLeNet and ResNet50 and set the embedding size as the same as other methods. RKD-DA is used for training student models.
The results are summarized in Table~\ref{tab:sota_metriclearning} where 
our method outperforms all the other methods on CUB-200-2011 regardless of backbone networks.
Among those using ResNet50, our method achieves the best performance on all the benchmark datasets.
Among those using GoogLeNet, our method achieves the second-best performance on Car 196 and Stanford Online Products, which is right below ABE8~\cite{Kim_2018_ECCV}.
Note that ABE8~\cite{Kim_2018_ECCV} requires additional multiple attention modules for each branches whereas ours is GoogLeNet with a single embedding layer.

\vspace{-0.3cm}
\subsubsection{Discussion}

\noindent\textbf{RKD performing better without $\ell2$ normalization.} One benefit of RKD over Triplet is that the student model is stably trained without $\ell2$ normalization. 
$\ell2$ norm forces output points of an embedding model to lie on the surface of unit-hypersphere, and thus a student model without $\ell2$ norm is able to fully utilize the embedding space. This allows RKD to better perform as shown in Table~\ref{tab:recall_small_dim}.
Note that DarkRank contains the triplet loss that is well known to be fragile without $\ell2$ norm. For example, ResNet18-128 trained with DarkRank achieves recall@1 of 52.92 without $\ell2$ norm (vs. 77.00 with $\ell2$ norm) on Cars 196. 

\smallbreak
\noindent\textbf{Students excelling teachers.} The similar effect has also been reported in classification~\cite{bagherinezhad2018label,furlanello2018born,Yim2017AGF}. The work of~\cite{bagherinezhad2018label,furlanello2018born} explains that the soft output of class distribution from the teacher may carry additional information, \eg, cross-category relationships, which cannot be encoded in one-hot vectors of ground-truth labels.
Continuous target labels of RKD (\eg, distance or angle) may also carry useful information, which cannot properly be encoded in binary (positive/negative) ground-truth labels used in conventional losses, \ie, the triplet loss. 

\begin{figure}
    \centering
    \begin{adjustbox}{minipage=\linewidth,scale=1.0}
        \centering
        \includegraphics[width=0.8\linewidth]{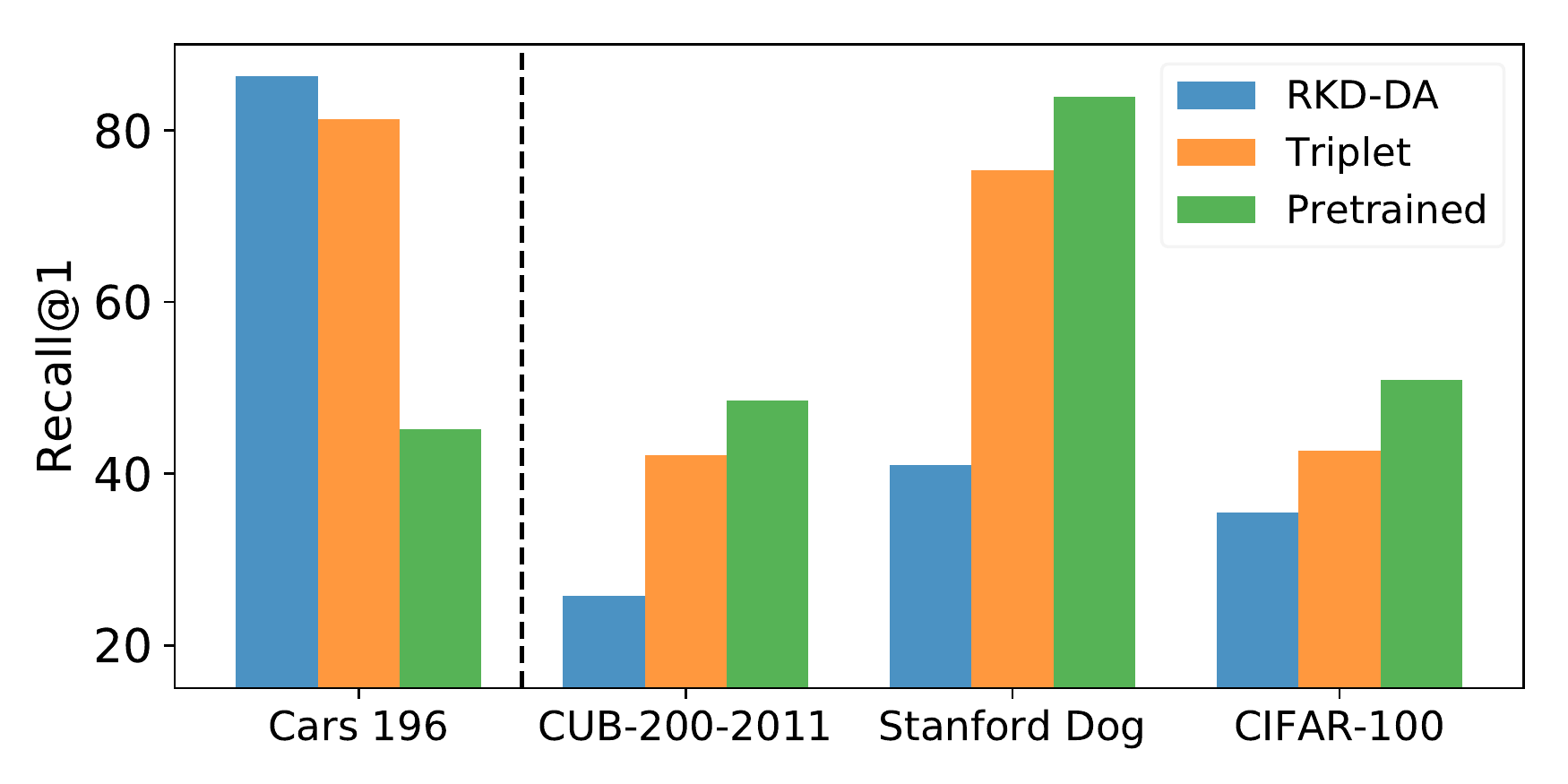}
    \vspace{-0.2cm}
    \caption{Recall@1 on the test split of Cars 196, CUB-200-2011, Stanford Dog and CIFAR-100. Both Triplet (teacher) and RKD-DA (student) are trained on Cars 196. The left side of the dashed line shows results on the training domain, while the right side presents results on other domains.}
    \label{fig:domain_adapt}
    \end{adjustbox}
    \vspace{-0.15cm}
\end{figure}

\smallbreak
\noindent\textbf{RKD as a training domain adaptation.}
Both Cars 196 and CUB-200-2011 datasets are originally designed for fine-grained classification, which is challenging due to severe intra-class variations and inter-class similarity.
For such datasets, effective adaptation to specific characteristics of the domain may be crucial; recent methods for fine-grained classification focus on localizing discriminative parts of target-domain objects \cite{peng2018object,yang2018learning,zhang2016picking}.
To measure the degree of adaptation of a model trained with RKD losses, 
we compare recall@1 on a training data domain with those on different data domains. 
Figure \ref{fig:domain_adapt} shows the recall@1 results on different datasets using a student model trained on Cars 196.
The student (RKD) has much lower recall@1 on different domains while the recall@1 of the teacher (Triplet) remains similarly to pretrained feature (an initial model). 
These results reveal an interesting effect of RKD that it strongly adapts models on the training domain at the cost of sacrificing generalization to other domains. 



\begin{figure*}
\includegraphics[width=0.7\textwidth]{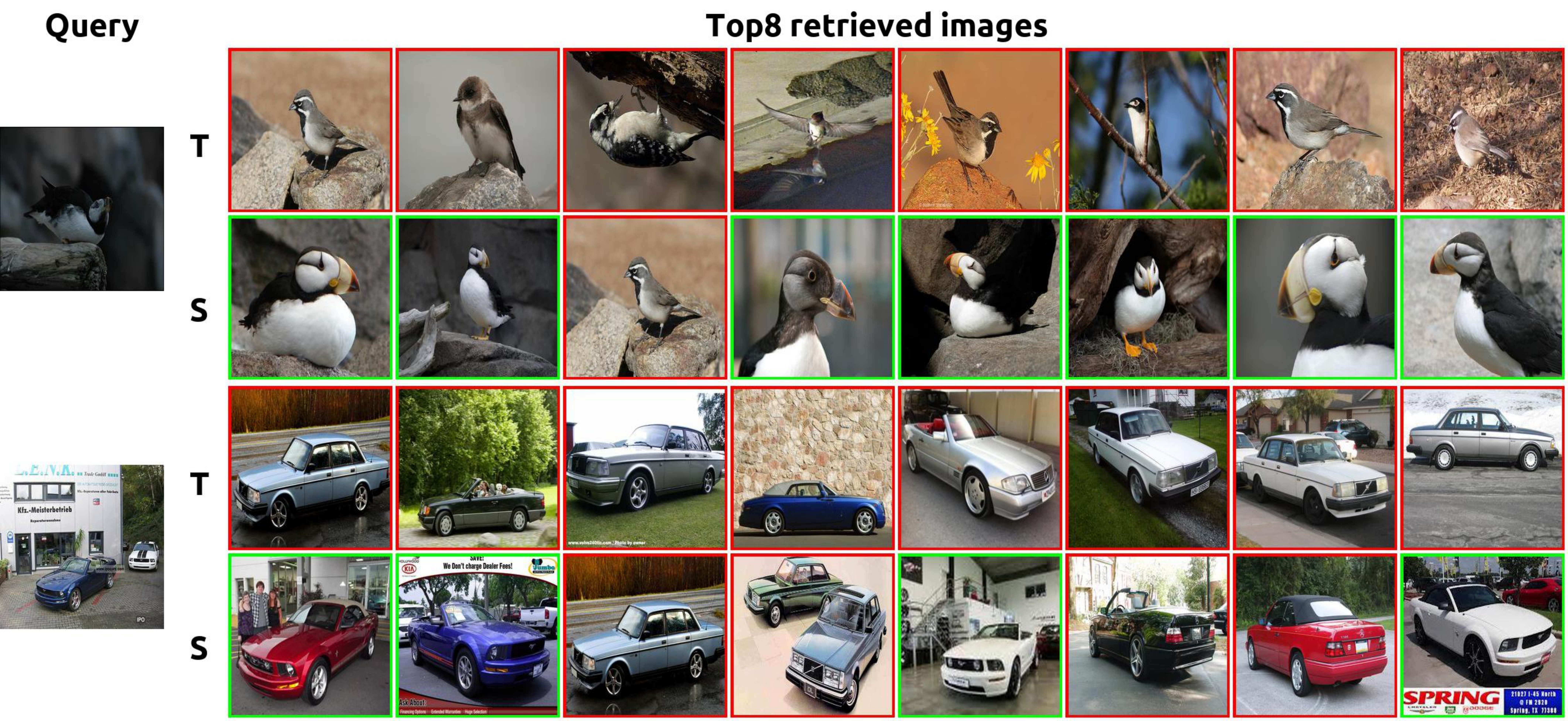}
\centering
\caption{Retrieval results on CUB-200-2011 and Cars 196 datasets. The top eight images are placed from left to right. Green and red bounding boxes represent positive and negative images, respectively. \textbf{T} denotes the teacher trained with the triplet loss while  \textbf{S} is the student  trained with RKD-DA. For these examples, the student gives better results than the teacher. }
\label{fig:qualitative_result}
\vspace{-0.3cm}
\end{figure*}

\subsection{Image classification}


\begin{table}[t!]
  \begin{center}
    \caption{Accuracy (\%) on CIFAR-100 and Tiny ImageNet.}
    \label{tab:classification_cifar}
    \vspace{-0.2cm}
    \small
    \begin{tabular}{c|c|c} 
       & CIFAR-100~\cite{Krizhevsky09} & Tiny ImageNet~\cite{tiny_imagenet} \\
      \hline
      Baseline & 71.26 & 54.45 \\
      \hdashline
      RKD-D & 72.27 & 54.97 \\
      RKD-DA  & 72.97 & 56.36\\
      \hdashline
      HKD~\cite{hinton2015distilling} & 74.26 & 57.65 \\
      HKD+RKD-DA & \textbf{74.66} & \textbf{58.15} \\
      \hdashline
      FitNet~\cite{romero2014fitnets} & 70.81 & 55.59 \\
      FitNet+RKD-DA & 72.98 & 55.54 \\
      \hdashline
      Attention~\cite{zagoruyko2016paying} & 72.68 & 55.51 \\
      Attention+RKD-DA & 73.53 & 56.55 \\
    \hline
    \hline
      Teacher & 77.76 & 61.55 \\
    \end{tabular}
  \end{center}
  \vspace{-0.7cm}
\end{table}

We also validate the proposed method on the task of image classification by comparing RKD with IKD methods, \eg, HKD~\cite{hinton2015distilling}, FitNet~\cite{romero2014fitnets} and Attention~\cite{zagoruyko2016paying}. 
We conduct experiments on CIFAR-100 and Tiny ImageNet datasets.
CIFAR-100 contains $32 \times 32$ sized images with 100 object categories, and Tiny ImageNet contains $64 \times 64$ sized images with 200 classes. 
For both datasets, we apply FitNet and Attention on the output of the second, the third, and the fourth blocks of CNN, and set $\lambda_\text{Attention} = 50$.
HKD is applied on the final classification layer on the teacher and the student, and we set temperature $\tau$ of HKD to be 4 and $\lambda_\text{HKD}$ to be 16 as in \cite{hinton2015distilling}.
RKD-D and RKD-A are applied on the last pooling layer of the teacher and the student, as they produce the final embedding before classification. We set $\lambda_\text{RKD-D} = 25$ and $\lambda_\text{RKD-A} = 50$.
For all the settings, we use the cross-entropy loss at the final loss in addition.
For both the teacher and the student, we remove fully-connected layer(s) after the final pooling layer and append a single fully-connected layer as a classifier.

For CIFAR-100, we randomly crop $32 \times 32$ images from zero-padded $40 \times 40$ images, and apply random horizontal flipping for data augmentation. We optimize the model using SGD with  mini-batch size $128$, momentum $0.9$ and weight decay $5\times10^{-4}$. We train the network for 200 epochs, and the learning rate starts from 0.1 and is multiplied by 0.2 at 60, 120, 160 epochs.
We adopt ResNet50 for a teacher model, and VGG11~\cite{simonyan2014very} with batch normalization for a student model.

For Tiny ImageNet, we apply random rotation, color jittering, and horizontal flipping for data augmentation. We optimize the model using SGD with mini-batch 128 and momentum 0.9. We train the network for 300 epochs, and the learning rate starts from 0.1, and is multiplied by 0.2 at 60, 120, 160, 200, 250 epochs. 
We adopt ResNet101 for a teacher model and ResNet18 as a student model.

Table \ref{tab:classification_cifar} shows the results on CIFAR-100 and Tiny ImageNet.
On both datasets, RKD-DA combined with HKD outperforms all configurations. 
The overall results reveal that the proposed RKD method is complementary to other KD methods; the model further improves in most cases when RKD is combined with another KD method. 

\subsection{Few-shot learning}
Finally, we validate the proposed method on the task of few-shot learning, which aims to learn a classifier that generalizes to new unseen classes with only a few examples for each new class.
We conduct experiments on standard benchmarks for few-shot classification, which are Omniglot~\cite{omniglot} and \textit{mini}ImageNet \cite{vinyals2016matching}. 
We evaluate RKD using the prototypical networks~\cite{snell2017prototypical} that learn an embedding network such that classification is performed based on distance from given examples of new classes.
We follow the data augmentation and training procedure of the work of Snell~\etal~\cite{snell2017prototypical} and the splits suggested by Vinyals \etal \cite{vinyals2016matching}. 
As the prototypical networks build on shallow networks that consist of only 4 convolutional layers,
we use the same architecture for the student model and the teacher, \ie, self-distillation, rather than using a smaller student network.
We apply RKD, FitNet, and Attention on the final embedding output of the teacher and the student.
We set $\lambda_\text{RKD-D}=50$ and $\lambda_\text{RKD-A} = 100$.
When RKD-D and RKD-A are combined together, we divide the final loss by 2.
We set $\lambda_\text{Attention} = 10$. 
For all the settings, we add the prototypical loss at the final loss.
As the common evaluation protocol of \cite{snell2017prototypical} for few-shot classification, we compute accuracy by averaging over 1000 randomly generated episodes for Omniglot, and 600 randomly generated episodes for \textit{mini}ImageNet. The Omniglot results are summarized in Table~\ref{tab:omniglot} while the \textit{mini}ImageNet results are reported with 95\% confidence intervals in Table~\ref{tab:miniImageNet}. They show that our method consistently improves the student over the teacher. 

\begin{table}
  \begin{center}
    \caption{Accuracy (\%) on Omniglot  \cite{omniglot}.}
     \vspace{-0.2cm}
    \label{tab:omniglot}
    \small
    \begin{tabular}{l|cc|cc} 
       & \multicolumn{2}{c|}{5-Way Acc.} & \multicolumn{2}{c}{20-Way Acc.} \\
       & 1-Shot & 5-Shot & 1-Shot & 5-Shot \\
    \hline
     RKD-D & 98.58 & \textbf{99.65} & 95.45 & \textbf{98.72} \\
     RKD-DA & \textbf{98.64} & 99.64 & \textbf{95.52} & 98.67 \\
    \hline
    \hline
     Teacher & 98.55 & 99.56 & 95.11 & 98.68 \\
    \end{tabular}
  \vspace{-0.5cm}
  \end{center}
\end{table}

\begin{table}
  \begin{center}
    \caption{Accuracy (\%) on \textit{mini}ImageNet~\cite{vinyals2016matching}.}
    \label{tab:miniImageNet}
    \small
    \vspace{-0.2cm}
    \begin{tabular}{l|c|c} 
       & \quad 1-Shot \space 5-Way \quad & \quad 5-Shot \space 5-Way \quad\\
    \hline
     RKD-D &  $49.66 \pm 0.84$ & $67.07 \pm 0.67$ \\
     RKD-DA & $50.02 \pm 0.83$ & $\textbf{68.16} \pm 0.67$ \\
     \hdashline
     FitNet & $\textbf{50.38} \pm 0.81$ & $68.08 \pm 0.65$\\
     Attention & $34.67 \pm 0.65$ & $46.21 \pm 0.70$\\
    \hline
    \hline
     Teacher & $49.1 \pm 0.82$ & $66.87 \pm 0.66$ \\
    \end{tabular}
  \vspace{-0.5cm}
  \end{center}
\end{table}

\section{Conclusion}
We have demonstrated on different tasks and benchmarks that the proposed RKD effectively transfers knowledge using mutual relations of data examples. In particular for metric learning, RKD enables smaller students to even outperform their larger teachers. While the distance-wise and angle-wise distillation losses used in this work turn out to be simple yet effective, the RKD framework allows us to explore a variety of task-specific RKD losses with high-order potentials beyond the two instances. 
We believe that the RKD framework opens a door to a promising area of effective knowledge transfer with high-order relations.      
\vspace{-0.3cm}
\paragraph{Acknowledgement:} This work was supported by MSRA Collaborative Research Program and also by Basic Science Research Program and Next-Generation Information Computing Development Program through the National Research Foundation of Korea funded by the Ministry of Science, ICT (NRF-2017R1E1A1A01077999, NRF-2017M3C4A7069369). 

{\small
\bibliographystyle{ieee}
\bibliography{egbib}
}

\end{document}